%% file: main.tex
\documentclass{article}
\usepackage{microtype}
\usepackage{graphicx}
\usepackage{subfigure}
\usepackage{booktabs} %
\usepackage{makecell}
\usepackage{pifont}
\usepackage{soul}

\usepackage{hyperref}

\usepackage[accepted]{icml2025}

\usepackage{amsmath}
\usepackage{amssymb}
\usepackage{mathtools}
\usepackage{amsthm}
\usepackage{multirow}
\usepackage{twemojis}

\usepackage[capitalize,noabbrev]{cleveref}

\theoremstyle{plain}

\theoremstyle{definition}

\theoremstyle{remark}

\definecolor{my_orange}{RGB}{255,152,0}
\definecolor{my_blue}{RGB}{33,150,243}

\usepackage[textsize=tiny]{todonotes}

\icmltitlerunning{The Best of Both Worlds: Integrating Language Models and Diffusion Models for Video Generation}

\begin{document}

\twocolumn[
\icmltitle{The Best of Both Worlds: Integrating Language Models and Diffusion Models for Video Generation}

\icmlsetsymbol{equal}{*}

\begin{icmlauthorlist}
\icmlauthor{Aoxiong Yin}{equal,yyy}
\icmlauthor{Kai Shen}{equal,comp}
\icmlauthor{Yichong Leng}{comp}
\icmlauthor{Xu Tan}{comp}
\icmlauthor{ Xinyu Zhou}{comp}
\icmlauthor{ Juncheng Li}{yyy}
\icmlauthor{Siliang Tang}{yyy}
\end{icmlauthorlist}

\icmlaffiliation{yyy}{College of Computer Science and Technology, Zhejiang University, Hangzhou, China}
\icmlaffiliation{comp}{Moonshot AI, Beijing, China}

\icmlcorrespondingauthor{Xu Tan}{tanxu@moonshot.cn} %
\icmlcorrespondingauthor{Juncheng Li}{junchengli@zju.edu.cn}

\vspace{0.2cm}
\begin{center}
\url{https://github.com/LanDiff/LanDiff}
\end{center}

\icmlkeywords{Machine Learning, ICML}

\vskip 0.3in
]

\printAffiliationsAndNotice{\icmlEqualContribution} %

\input{sec/abstract}
\input{sec/intro.tex}
\input{sec/method.tex}

\input{sec/exp.tex}

\input{sec/related_work.tex}
\input{sec/conclusion.tex}

\clearpage
\bibliography{icml_2025_ref,manual}
\bibliographystyle{icml2025}

\newpage
\appendix
\onecolumn

\input{sec/appendix}

\end{document}

%% file: sec/abstract.tex
\begin{abstract}
Recent advancements in text-to-video (T2V) generation have been driven by two competing paradigms: autoregressive language models and diffusion models. However, each paradigm has intrinsic limitations: language models struggle with visual quality and error accumulation, while diffusion models lack semantic understanding and causal modeling. In this work, we propose LanDiff, a hybrid framework that synergizes the strengths of both paradigms through coarse-to-fine generation. Our architecture introduces three key innovations: (1) a semantic tokenizer that compresses 3D visual features into compact 1D discrete representations through efficient semantic compression, achieving a $\sim$14,000$\times$ compression ratio; (2) a language model that generates semantic tokens with high-level semantic relationships; (3) a streaming diffusion model that refines coarse semantics into high-fidelity videos. Experiments show that LanDiff, a 5B model, achieves a score of 85.43 on the VBench T2V benchmark, surpassing the state-of-the-art open-source models Hunyuan Video (13B) and other commercial models such as Sora, Kling, and Hailuo. Furthermore, our model also achieves state-of-the-art performance in long video generation, surpassing other open-source models in this field. Our demo can be viewed at \url{https://landiff.github.io/}.
\end{abstract}

%% file: sec/intro.tex
\section{Introduction}
Text-to-video (T2V) \cite{blattmann_StableVideoDiffusion_2023, kondratyuk_VideoPoetLargeLanguage_2024, wang_Emu3NextTokenPrediction_2024, yang_CogVideoXTexttoVideoDiffusion_2024} has made significant progress in recent years, becoming an important research direction in the fields of computer vision and artificial intelligence.
Recent works in T2V models have primarily revolved around two predominant paradigms: autoregressive large language model (LLM)-based \cite{kondratyuk_VideoPoetLargeLanguage_2024, wang_Emu3NextTokenPrediction_2024} frameworks and diffusion-based architectures \cite{blattmann_StableVideoDiffusion_2023, yang_CogVideoXTexttoVideoDiffusion_2024}.
However, each paradigm has their own advantages and limitations, as shown in \autoref{tab:ad_and_dis}.

\begin{figure}[ tbp]
    \centering
    \includegraphics[width=\linewidth]{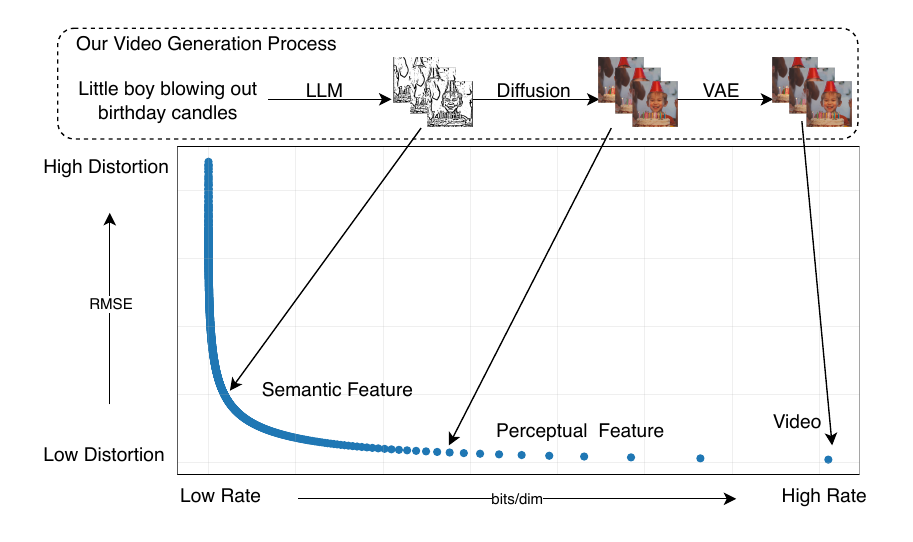}
    \vspace{-3mm}
    \caption{The rate-distortion curve illustrates how visual distortion decreases as the number of transmitted bits increases. With just a small number of bits representing high-level semantic features, we can already achieve relatively low visual distortion. Building on this information-theoretic insight, LanDiff combines the strengths of both paradigms: LLMs efficiently generate compact semantic features in the first stage, followed by diffusion models that add perceptual details in the second stage, before final decoding to pixels via VAE. Data from \citet{ddpm}, illustration is conceptual.}
    \label{fig:fig1}
    \vspace{-3mm}
\end{figure}

\textbf{From a representation perspective}, LLM-based methods leverage discrete tokenization to explicitly encode high-level semantics through vector quantization, effectively prioritizing conceptual abstraction and narrative coherence. However, this discretization inherently sacrifices low-level visual fidelity due to information compression, resulting in low reconstruction quality. In contrast, diffusion-based approaches employ continuous latent representations to preserve much more perceptual details, enabling superior reconstruction quality at the cost of diluted semantic interpretability, as hierarchical features remain entangled in the latent space. 
\textbf{From a generative modeling perspective}, LLM-based systems adopt autoregressive modeling to enforce causal dependencies between video frames, ensuring strong temporal coherence. However, the autoregressive generation inherently risks error propagation across time steps, where inaccuracies in early predictions amplify during decoding. In contrast, diffusion-based methods employ non-autoregressive generation, refining outputs in parallel through iterative denoising steps. Although this design mitigates error accumulation and enhances generation flexibility, the absence of explicit causal constraints often leads to temporal inconsistencies or semantic hallucinations.

\begin{table}[tbp]
    \centering
    \caption{The comparison between LLM based and diffusion based T2V systems. The advantages and disadvantages are marked by {\large\texttwemoji{smiley}} and {\large\texttwemoji{slightly_frowning_face}}, respectively.}
    \label{tab:ad_and_dis}
    \resizebox{0.99\linewidth}{!}{
    \begin{tabular}{lll} %
        \toprule
    Methods          & Representations        & Modeling  \\  %
    \midrule
    LLM   & \makecell[l]{Discrete Tokens\\Low Reconstruction Quality {\Large\texttwemoji{slightly_frowning_face}}\\Highlight Semantic Information {\Large\texttwemoji{smiley}}}       & \makecell[l]{Autoregressvie\\No Refinement {\Large\texttwemoji{slightly_frowning_face}} \\ Causal Modeling {\Large\texttwemoji{smiley}}}    \\
    \midrule
    Diffusion   &  \makecell[l]{Continuous Vectors\\ High Reconstruction Quality {\Large\texttwemoji{smiley}} \\ Lack Semantic Information {\Large\texttwemoji{slightly_frowning_face}}}    & \makecell[l]{Non-Autoregressvie\\Progressive Refinement {\Large\texttwemoji{smiley}} \\ Non-causal Modeling {\Large\texttwemoji{slightly_frowning_face}}}    \\
    \bottomrule
    \end{tabular}
    }
    \vspace{-2mm}
\end{table}

In this work, we propose a hybrid architecture that synergizes the strengths of both \textbf{Lan}guage models and \textbf{Diff}usion models, named LanDiff, through a coarse-to-fine generation paradigm. As shown in \autoref{fig:fig1}, inspired by the human creation of video which will generate the high-level storyline first and then add low-level visual details based on the storyline to form the video, we design a two-stage video generation process with the number of bits gradually increasing and carefully design the autoregressive model and the diffusion model to be responsible for different stages of T2V generation, so as to play to their strengths and avoid their weaknesses. In detail, 1) at the low-bit position ``semantic feature'', the low-bit information ensures that the token sequence is not too long, and the high-level information makes it easier for the model to capture the overall semantic entity motion of the video, so as to play to the strengths of the autoregressive model and avoid its weaknesses. Thus we use LLM to generate a coarse-grained video in the first stage; 2) at the high-bit position ``perceptual feature'', since we have already obtained the coarse-grained with rich semantic and time-serial information, we only need to focus on how to add details to the coarse-grained video. Thus we apply a diffusion model in the second stage. 
Finally, a VAE decoder transforms the generated ``perceptual feature'' into the final RGB video output.
By unifying these complementary mechanisms, we demonstrate that hybrid architectures can overcome the inherent limitations of isolated approaches, enabling coherent, semantically faithful, and visually compelling video generation from textual descriptions, as shown in \autoref{tab:cmp}.

\begin{table}[tbp]
    \centering
    \caption{The comparison between LanDiff and previous large-scale T2V systems. The compression rates of LLM-based and diffusion-based models are illustrated using VideoPoet \cite{kondratyuk_VideoPoetLargeLanguage_2024} and CogVideoX \cite{yang_CogVideoXTexttoVideoDiffusion_2024} as examples, respectively.}
    \label{tab:cmp}
    \resizebox{0.99\linewidth}{!}{
    \begin{tabular}{llll} %
        \toprule
    Models          & LLM        & Diffusion      & LanDiff   \\  %
    \midrule
   Highlight Semantic Information? & \ding{51} & \ding{55} & \ding{51} \\
   Progressive Refinement? & \ding{55}       & \ding{51}   & \ding{51}\\
    Causal Modeling?   & \ding{51}         & \ding{55} & \ding{51}   \\
    High Visual Quality?   & \ding{55}       & \ding{51}   & \ding{51}    \\
    \midrule
    Compression Ratio $\uparrow$   & $\sim $256       & $\sim $1024   & $\sim $14000    \\
    Long video Generation?   & \ding{55}       & \ding{55}   & \ding{51}    \\
    \bottomrule
    \end{tabular}
    }
    \vspace{-3mm}
    \end{table}

With this design, the ideal semantic feature should contain high-level semantic information and motion information and only require a few bits to represent.
We achieve this goal by performing extreme compression on video representations rich in high-level semantics.
For video representation, we select the Theia model \cite{shang_TheiaDistillingDiverse_2024} as our visual representation backbone, which has been distilled from multiple visual understanding and self-supervised representation models, ensuring the encoded features contain rich semantic information.
To achieve extreme compression and reduce the number of bits, we design an efficient tokenizer to compress 3D visual features into 1D discrete representations.
The tokenizer is based on the Transformer~\cite{vaswaniAttentionAllYou2017} structure, uses query embedding to aggregate visual features, and has a higher compression rate than CNN-based structures \cite{yu_ImageWorth32_2024}.
To further compress the video by fully utilizing the temporal redundancy of the video, inspired by the MP4 video encoding algorithm \cite{le1991mpeg}, we divide the video into keyframes and non-keyframes, and set more numbers of tokens for keyframes.
The detailed design is shown in \autoref{sec:tokenizer}. 
For the diffusion model, we use generated semantic tokens as conditions and generate the target video by gradually removing the noises. To better support the long-video generation, we train a chunk-wise streaming diffusion model that only uses a limited number of historical frames as conditions, thereby greatly reducing the computational cost of training and inference.

Thanks to these designs, our LanDiff has made significant progress in spatial relationship compliance, action coherence, visual quality, etc.
Specifically, 1) for short video generation, our 5B model achieved a score of 85.43 on the VBench T2V benchmark, surpassing the state-of-the-art open-source models Hunyuan Video (13B) and other commercial models such as Sora, Kling, and Hailuo;
2) for long video generation, after testing by the VBench T2V benchmark, our model also achieved state-of-the-art performance, surpassing other open-source models in this field.
Our video examples can be viewed at \url{https://landiff.github.io/}.

%% file: sec/method.tex
\begin{figure*}[tbp]
    \centering
    \includegraphics[width=0.8\textwidth]{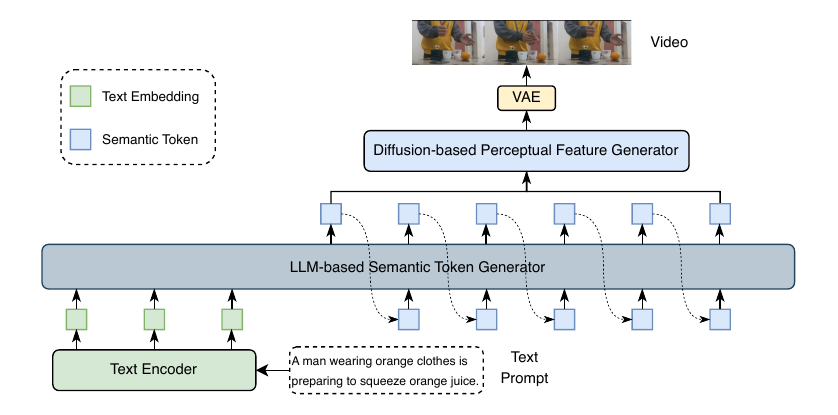}
    \caption{The architecture of LanDiff. Given text inputs, we first extract text embeddings and employ an LLM to generate semantic tokens in the first stage. Subsequently, we utilize a diffusion model to synthesize perceptual features conditioned on these semantic tokens, followed by a VAE decoder that transforms these features into the final video frames.}
    \label{fig:model}
\end{figure*}

\section{Method}
\label{sec:method}
In this work, we propose a novel text-to-video generation framework that synergistically integrates the strengths of autoregressive modeling and diffusion processes while circumventing their respective limitations.
The framework mainly consists of the following components: 1) an efficient tokenizer that transforms 3D visual features into compact 1D discrete representations while preserving and enhancing their semantic information; 2) a language model that performs temporal sequence modeling to generate semantic tokens representing video blueprints from textual descriptions; 3) a streaming diffusion model that progressively refines coarse semantic videos by adding fine-grained details to produce high-quality VAE features; and 4) a VAE decoder that reconstructs the final video frames from the refined VAE features.

\subsection{Video Semantic Tokenizer}
In this section, we introduce a novel video semantic tokenizer that efficiently compresses the video into semantic information. 
Firstly, we will introduce the video semantic representation used in tokenization. 
Then, considering the high redundancy of video in both spatial and temporal dimensions, we introduce two compression strategies: 1) a query-based causal tokenization that efficiently reduces spatial redundancy while preserving essential semantic information through vector quantization; 2) inspired by MP4~\cite{le1991mpeg}, we implement video frame grouping to minimize temporal redundancy by treating grouped frames as a unit, where the first frame (I-frame) is fully encoded while subsequent frames (P-frames) only capture the temporal changes by referencing content from previous frames.

\label{sec:tokenizer}
\textbf{Video Semantic Representation.}
Generally, video representations can be divided into two categories: 1) some methods~\cite{wang_Emu3NextTokenPrediction_2024, yu_LanguageModelBeats_2024} directly utilize an autoencoder to learn the video representations. 2) 
Some works \cite{koh_GeneratingImagesMultimodal_2023,jin_UnifiedLanguageVisionPretraining_2023,sun_GenerativeMultimodalModels_2024,jin_VideoLaVITUnifiedVideoLanguage_2024} use pre-trained visual self-supervised learning features (SSL) as video representations. 
Compared with the first directly learned autoencoder latents which contain lots of visual details, the second SSL features maintain much more semantic features, which enable the LLM to focus more on the high-level semantic information of the video. 

Based on these thoughts, we choose the pretrained SSL features as the video representations. For better usage of different SSL models, we choose Theia model \cite{shang_TheiaDistillingDiverse_2024}, which is a unified visual feature extractor distilled from multiple visual task models, including CLIP \cite{radfordLearningTransferableVisual2021}, SAM \cite{kirillov_SegmentAnything_2023}, DINOv2 \cite{oquab_DINOv2LearningRobust_2024}, ViT~\cite{dosovitskiy_ImageWorth16x16_2021}, and Depth-Anything \cite{yang_DepthAnythingUnleashing_2024}.

\begin{figure*}[tbp]
    \centering
    \includegraphics[width=0.75\textwidth]{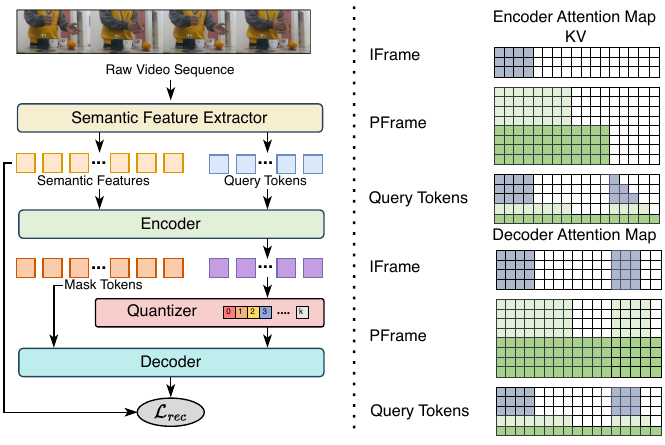}
    \caption{Proposed architecture of the video semantic tokenizer. We use query tokens to compress the semantic sequence length. Furthermore, we group the frames into groups (3 frames in a group in this figure). In a group, the first frame is the IFrame and the rest frames are PFrames. We use different query token numbers for them. The attention mask design is shown in the right.}
    \vspace{-1mm}
    \label{fig:tokenizer}
\end{figure*}

\textbf{Tokenizer Design.}
In this part, we introduce the design of our tokenizer, which leverages the query tokens to compress the video semantic features and uses quantization to discrete the video semantic representation while minimizing the reconstruction loss.

In detail, firstly, we extract the semantic features using the Theia model and flatten it to obtain $F \in \mathbb{R}^{(T \times H \times W) \times D}$. Inspired by TiTok~\cite{yu_ImageWorth32_2024}, we use N randomly initialized tokens as query tokens $Q\in \mathbb{R}^{N \times D}$ and concatenate them with the semantic features $F$. Then we use a transformer encoder to encode them and only take the encoded features of the query tokens.
\begin{equation}
Z_{Q}=\operatorname{Enc}([F ; Q]),
\end{equation}
where $[;]$ represents the concatenation operation and $Z_{Q} \in \mathbb{R}^{N \times D}$. We then apply vector quantization on $Z_{Q}$ by train a VQ-VAE model and obtain the quantized feature $\hat{Z}_{Q}$. In the decoding stage, the quantized feature $\hat{Z}_{Q}$ is used as a condition, and then a sequence of mask tokens $M \in \mathbb{R}^{T \times H \times W}$ are added in front of the $\hat{Z}_{Q}$ to form the inputs of decoder. Then we only take the features of the mask tokens as follows:
\begin{equation}
    \hat{F}=\operatorname{Dec}([M;\hat{Z}_{Q}]),
\end{equation}
where $\hat{F} \in \mathbb{R}^{(T \times H \times W)\times D}$ represents the reconstructed feature, and $M$ represents the mask tokens. Inspired by \citet{wang2024maskgct,huang2023repcodec}, we minimize the reconstruction loss of the video semantic feature during training the VQ-VAE.

For the VQ-VAE, we follow the method of \citet{yu_VectorquantizedImageModeling_2022,wang2024maskgct}. 
We update the codebook using exponential moving average (EMA). 
We also optimize the codebook with the video semantic feature reconstruction loss. The loss function is shown as follows:
\begin{equation}
    \mathcal{L} = \lambda_{rec} || \hat{F} - F ||_{2} + \lambda_{commit}|| \text{sg}(\hat{Z}_{Q}) - Z_{Q} ||_{2},
\end{equation}
where $\text{sg}()$ is the stop-gradient operation, $||\cdot||_2$ is the L2 loss.

\textbf{Video Frame Grouping.}
For video, a straightforward observation is the redundancy in time series (i.e., the difference between adjacent video frames is minimal). Intuitively, we can achieve a better compression rate by modeling the difference between adjacent video frames instead of tokenizing all frames equally. Inspired by the popular video compression method MP4~\cite{le1991mpeg}, given $N$ frames of a video, we first split them into $N / T$ groups (i.e., $T$ frames as a group. For clarification, as shown in \autoref{fig:tokenizer}, we use $T=3$ as an example.). Then: 1) we will model different groups independently; 2) for each group, we will fully encode the first frame (Intra-coded Frame, IFrame), while for the remaining $[1, T-1]$ frames (the Predictive-coded Frame, PFrame), we encode them by referencing their previous frames.

To achieve this, within a group, the first frame (IFrame) will only see itself and have a large number of query tokens (e.g., 3 query tokens in \autoref{fig:tokenizer}) to achieve better reconstruction quality. For the rest frame (PFrame) $i \in [1, T-1]$, it will see the previous frames (i.e., frames $j \in [0, i-1]$) and have a small number of query tokens (e.g., 1 query token which is $1/3$ of the IFrame) to force the model to learn the difference. 
Technically, as shown in the attention mask on the right side of \autoref{fig:tokenizer}, we use frame-level causal masks for the feature sequence during encoding.
The query tokens are also divided according to the frame, and each token can only attend to the features of the corresponding frame and the previous frames.
During decoding, the mask token corresponding to each frame can see the previous features, the corresponding query tokens, and the previous query tokens.

\subsection{Language Model for Semantic Token Generation}
\label{sec:lm}
As shown in \autoref{fig:model}, after training an efficient tokenizer, we use a language model to generate semantic tokens autoregressively based on text.
Specifically, we first use a pre-trained text encoder T5-XXL \cite{raffel_ExploringLimitsTransfer_2020} to extract text features $X$.
We use the video tokenizer encoder trained in \autoref{sec:tokenizer} to convert the video into a 1D discrete token sequence $Y$.
To enhance the controllability of the generated results, we introduce additional control conditions $CC$. It includes: 1) frames condition for requiring the model to generate videos with a specified number of frames; 2) motion score condition, which is a value between 0 and 1, used to control the degree of motion in the generated video.
The discrete tokens are converted into embedding vectors during generation and added with positional encoding, and then concatenated with the control conditions as input.
The structure of the language model follows the typical LLaMA model \cite{touvron_LLaMAOpenEfficient_2023} network structure.
We train the model from scratch, using cross-entropy loss as the loss function.

\begin{equation}
    \mathcal{L}_{\text {LM}}=\mathbb{E} [-\log p(Y_{i} | X, CC,Y_{<i})]
\end{equation}

\subsection{Diffusion Model for Perceptual Feature Generation}
\label{sec:detokenizer}
The video detokenizer in LanDiff is responsible for converting semantic tokens into VAE latent vectors.
As shown in \autoref{fig:model} we use a conditional diffusion model to complete this task. Especially to support long-video generation, we also design the streaming inference strategy. In this section, we will first introduce the architecture for a single semantic token chunk. Then we will introduce the chunk-wise streaming strategy. 

\begin{figure}[tbp]
    \centering
    \includegraphics[width=0.9\linewidth]{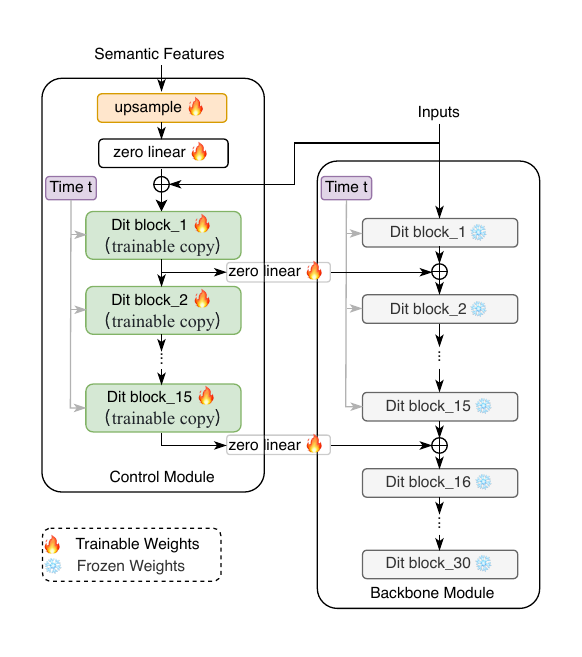}
    \caption{Proposed diffusion model structure. We use a ControlNet-style control module to guide the model to generate perceptual feature based on semantic features.}
    \label{fig:control}
\end{figure}

\textbf{Architecture.} 
Our diffusion model employs an architecture similar to MMDiT\cite{sd3,yang_CogVideoXTexttoVideoDiffusion_2024}.
Specifically, 1) we use the video tokenizer decoder trained in \autoref{sec:tokenizer} to decode the semantic tokens into semantic features $\hat{F}$. Then we use the semantic features $\hat{F}$ as a condition to guide the diffusion model to generate videos; 2) we inject the control signals into the model in a similar way to ControlNet \cite{zhang_AddingConditionalControl_2023}, as shown in \autoref{fig:control}. In detail, during training, the parameters of the main model are not updated, the control module copies the parameters of the first half layers of the main model, and adds them to the output of the main model after a linear layer initialized with zeros; 3) to make the semantic features match the target VAE features in the space dimension, we additionally add an upsampling module.

During training, given the input video chunk $V \in \mathbb{R}^{(T \times H \times W)\times D}$, where $T$ is the frame length of the VAE latent and $D$ is the VAE latent feature dimension, the diffusion algorithm progressively adds noise to the video and produces a noisy video $V^t$, where $t$ represents the number of times noise is added. The diffusion algorithm uses time step $t$, and semantic features $\hat{F} \in \mathbb{R}^{(T \times H \times W)\times D}$ as conditions, and then uses a network $\epsilon_\theta$ to predict the noise added to the noisy video $V_t$ through:
\begin{equation}
    \mathcal{L} = \mathbb{E}_{V, t, c_s, \epsilon \sim \mathcal{N}(0, 1) }\Big[ \Vert \epsilon - \epsilon_\theta(V^{t}, t, \hat{F})) \Vert_{2}^{2}\Big]
    \label{eq:diff_loss}
\end{equation}

\textbf{Chunk-wise Streaming Strategy.} To support long video generation scenario whose semantic token sequence is very long and difficult to generate as a whole, we propose the chunk-wise streaming strategy. During training, given video latent chunk $V \in \mathbb{R}^{(T \times H \times W)\times D}$, to maintain the appearance continuity in the video, we use the first half of $V$ ($V_{l}$) as the prompt and generate the second half of $V_{r}$. In detail, we do not add noise to the first half chunk and give time condition $t$ to $0.999$. We also randomly mask the first half chunk with a ratio of $20\%$.
The loss function for training at this time is:
\begin{equation}
    \mathcal{L} = \mathbb{E}_{V_{r}, t, c_s, \epsilon \sim \mathcal{N}(0, 1) }\Big[ \Vert \epsilon - \epsilon_\theta(V_{l},V_{r}^{t}, t, c_s)) \Vert_{2}^{2}\Big].
\end{equation}

During inference, the first $L/2$ VAE latents will be generated without the prompt. For the following VAE latents, we will accumulate $L/2$ tokens to form a chunk and use the previous chunk as prompt.

%% file: sec/exp.tex
\section{Experiments and Results}
\subsection{Experimental Settings}
\textbf{Datasets.}
For the video tokenizer and language model, we use an internal dataset with 200M video-text pairs for training.
All videos with a duration of less than 6s are filtered, and the videos are kept in the aspect ratio and then scaled to around 480x720 resolution for center cropping.
Consistent with CogVideoX \cite{yang_CogVideoXTexttoVideoDiffusion_2024}, we set the fps of all videos to 8.
For the diffusion model, we select a dataset containing 3M high-quality video-text pairs for training.
To evaluate the performance of the text-to-video generation model, we use the prompts provided by the widely used VBench \cite{huang_VBenchComprehensiveBenchmark_2024} T2V benchmark to generate videos.

\textbf{Implementation Details.}
We use interpolation of positional encoding to enable the encoder of the Theia \cite{shang_TheiaDistillingDiverse_2024} model to handle 480x720 resolution videos.
For the video tokenizer, we set the group size $T$ mentioned in \autoref{sec:tokenizer} to 13.
We set the number of tokens $I$ corresponding to the IFrame to 330, and the number of tokens $P$ corresponding to the PFrame to 74. 
On average, for a 480x720 resolution video of one second, our tokenizer generates about 200 tokens.
In contrast, common tokenizers such as MagViT2 \cite{yu_LanguageModelBeats_2024} generate about 10,000 tokens per second for videos of this resolution, and our sequence length is about 1/50 of MagViT2.
The dimension of the tokenizer's codebook is set to 16, and the vocabulary size is set to 2048.
For the language model, we use the LLaMA \cite{touvron_LLaMAOpenEfficient_2023} structure, with 2B model parameters, and use 1D RoPE \cite{su_RoFormerEnhancedTransformer_2024} positional encoding.
We set the text, motion score, and frames conditions in \autoref{sec:lm} to null with probabilities of 10\%, 50\%, and 50\%, respectively.
We apply classifier-free guidance \cite{ho_ClassifierFreeDiffusionGuidance_2022} for better generation quality, and the guidance scale is set to 6.5.
We do not use top-k and top-p sampling.
We use the 2B version of the CogVideoX\footnote{\url{https://huggingface.co/THUDM/CogVideoX-2b}} \cite{yang_CogVideoXTexttoVideoDiffusion_2024} model as the base model for our video detokenizer.
We copy the first 15 layers of the base model as the proposed trainable control module in \autoref{sec:detokenizer}.
We use a structure similar to the VQ-GAN \cite{esserTamingTransformersHighResolution2021} decoder as the upsampling module and change the upsampling method to pixelshuffle \cite{shi_RealtimeSingleImage_2016}.
The total number of parameters of the video detokenizer is 3B, and the number of parameters of the trainable control module is 1B.
During inference, we follow the same sampling strategy as \citet{yang_CogVideoXTexttoVideoDiffusion_2024}.
We list the structural settings and training details of each module in the appendix.

\textbf{Evaluation Metrics.}
To evaluate the text-to-video generation task, we use the metrics proposed in the VBench and VBench-Long \cite{huang_VBenchComprehensiveBenchmark_2024} benchmarks.

\textbf{Evaluation Baselines.}
We compare LanDiff with baselines: 1) Sora \cite{Sora}. 2) Jimeng \cite{Jimeng}. 3) Hailuo \cite{Hailuo}. 4) OpenSoraPlan V1.1 \cite{lin_OpenSoraPlanOpenSource_2024}. 5) Kling \cite{VideoFusion}. 6) InstructVideo \cite{yuan_InstructVideoInstructingVideo_2024}. 7) Gen-3 \cite{Gen-3}. 9) Latte-1 \cite{ma_LatteLatentDiffusion_2024} 9) HiGen \cite{higen}. 10) AnimateDiff-V2 \cite{guo_AnimateDiffAnimateYour_2024}. 11) Show-1 \cite{zhang_Show1MarryingPixel_2023}. 12) Pika \cite{Pika}. 13) VideoCrafter-2.0 \cite{chen_VideoCrafter2OvercomingData_2024}. 14) OpenSora V1.2 \cite{opensora}. 15) LTX-Video \cite{hacohen_LTXVideoRealtimeVideo_2024}. 16) Mochi-1 \cite{genmo2024mochi}. 17) CogVideoX \cite{yang_CogVideoXTexttoVideoDiffusion_2024}. 18) Vchitect-2.0 \cite{fan_Vchitect20ParallelTransformer_2025}. 19) RepVideo \cite{si_RepVideoRethinkingCrossLayer_2025}. 20) HunyuanVideo \cite{kong_HunyuanVideoSystematicFramework_2024}. 21) ARLON \cite{li_ARLONBoostingDiffusion_2024}. 22) Emu3 \cite{wang_Emu3NextTokenPrediction_2024}.

\subsection{Experimental Results}

\begin{table*}[tbp]
    \caption{\textbf{Performance comparison of Text-to-video (T2V) generation between our LanDiff and other state-of-the-art models on VBench benchmark.} We selected 8 out of the 16 evaluation dimensions from VBench, along with Total Score, Quality Score, and Semantic Score, for presentation. The complete results with all 16 evaluation dimensions can be found in the appendix \autoref{tab:t2v_left}. The best and second-best scores are highlighted in \textbf{bold} and \underline{underline}, respectively. $\dagger$ indicates the scores we reproduced, while $\ddagger$ indicates the scores from the original papers, and other scores are from the VBench benchmark.}
    \label{tab:t2v}
        \small
    \setlength\tabcolsep{2pt}
    \centering
    \resizebox{0.9\linewidth}{!}{
    \begin{tabular}{lccccccccccccc}
        \toprule
        \thead{\textbf{Model} \\ \textbf{Name}} & \thead{\textbf{Type}} & \thead{\textbf{Model} \\ \textbf{Size}} & \thead{\textbf{Total} \\ \textbf{Score}} & \thead{\textbf{Quality} \\ \textbf{Score}} & \thead{\textbf{Semantic} \\ \textbf{Score}} & \thead{\textbf{background} \\ \textbf{consistency}} & \thead{\textbf{dynamic} \\ \textbf{degree}} & \thead{\textbf{motion} \\ \textbf{smoothness}} & \thead{\textbf{multiple} \\ \textbf{objects}} & \thead{\textbf{object} \\ \textbf{class}} & \thead{\textbf{scene}} & \thead{\textbf{spatial} \\ \textbf{relationship}} & \thead{\textbf{subject} \\ \textbf{consistency}} \\
        \midrule
        \multicolumn{14}{c}{\textit{Open Sourced Models}} \\
        \midrule
        InstructVideo & Diffusion & 1.3B & 76.61 & 81.56 & 56.81 & 96.97 & 69.72 & 96.62 & 21.57 & 73.26 & 22.21 & 43.49 & 95.30 \\
        Latte-1 & Diffusion & 0.7B & 77.29 & 79.72 & 67.58 & 95.40 & 68.89 & 94.63 & 34.53 & 86.53 & 36.26 & 41.53 & 88.88 \\
        OpenSoraPlan V1.1 & Diffusion & 2.7B & 78.00 & 80.91 & 66.38 & 96.73 & 47.72 & 98.28 & 40.35 & 76.30 & 27.17 & 53.11 & 95.73 \\
        Show-1 & Diffusion & 6.3B & 78.93 & 80.42 & 72.98 & 98.02 & 44.44 & 98.24 & 45.47 & 93.07 & 47.03 & 53.50 & 95.53 \\
        OpenSora V1.2 & Diffusion & 1.1B & 79.76 & 81.35 & 73.39 & 97.61 & 42.39 & 98.50 & 51.83 & 82.22 & 42.44 & 68.56 & 96.75 \\
        LTX-Video & Diffusion & 1.9B & 80.00 & 82.30 & 70.79 & 97.20 & 54.35 & 98.96 & 45.43 & 83.45 & 51.07 & 65.43 & 96.56 \\
        Mochi-1 & Diffusion & 10B & 80.13 & 82.64 & 70.08 & 97.28 & 61.85 & 99.02 & 50.47 & 86.51 & 36.99 & 69.24 & 96.99 \\
        AnimateDiff-V2 & Diffusion & 1.3B & 80.27 & 82.90 & 69.75 & 97.68 & 40.83 & 97.76 & 36.88 & 90.90 & 50.19 & 34.60 & 95.30 \\
        VideoCrafter-2.0 & Diffusion & 1.7B & 80.44 & 82.20 & 73.42 & 98.22 & 42.50 & 97.73 & 40.66 & 92.55 & 55.29 & 35.86 & 96.85 \\
        CogVideoX-2B & Diffusion & 2B & 80.91 & 82.18 & 75.83 & 96.63 & 59.86 & 97.73 & 62.63 & 83.37 & 51.14 & 69.90 & 96.78 \\
        Emu3 $\ddagger$ & LLM & 8B & 80.96 & N/A & N/A & 97.69 & 79.27 & 98.93 & 44.64 & 86.17 & 37.11 & 68.73 & 95.32 \\
        Vchitect-2.0-2B & Diffusion & 2B & 81.57 & 82.51 & 77.79 & 96.53 & 58.33 & 97.76 & 69.35 & 87.81 & \textbf{57.51} & 54.64 & 96.42 \\
        CogVideoX-5B & Diffusion & 5B & 81.61 & 82.75 & 77.04 & 96.52 & 70.97 & 96.92 & 62.11 & 85.23 & 53.20 & 66.35 & 96.23 \\
        DiT $\dagger$ & Diffusion & 7B & 81.85 & 82.70 & 78.42 & 97.65 & 51.02 & 98.94 & 69.63 & 93.40 & 57.10 & 60.89 & 97.05 \\
        RepVideo & Diffusion & 2B & 81.94 & 82.70 & 78.91 & 96.56 & 57.78 & 98.13 & 71.18 & 87.83 & 52.96 & 74.74 & 96.25 \\
        Vchitect-2.0[E] & Diffusion & 2B & 82.24 & 83.54 & 77.06 & 96.66 & 63.89 & 98.98 & 68.84 & 86.61 & 56.57 & 57.55 & 96.83 \\
        HunyuanVideo & Diffusion & 13B & 83.24 & 85.09 & 75.82 & 97.76 & 70.83 & 98.99 & 68.55 & 86.10 & 53.88 & 68.68 & 97.37 \\
        \midrule
        \multicolumn{14}{c}{\textit{Close Sourced Models}} \\
        \midrule
        Pika-1.0 & Diffusion & N/A & 80.69 & 82.92 & 71.77 & 97.36 & 47.50 & \textbf{99.50} & 43.08 & 88.72 & 49.83 & 61.03 & 96.94 \\
        Kling & Diffusion & N/A & 81.85 & 83.39 & 75.68 & 97.60 & 46.94 & \underline{99.40} & 68.05 & 87.24 & 50.86 & 73.03 & \textbf{98.33} \\
        Jimeng & Diffusion & N/A & 81.97 & 83.29 & 76.69 & 98.39 & 38.43 & 98.09 & 69.08 & 89.62 & 44.94 & \textbf{77.45} & 97.25 \\
        Gen-3 & Diffusion & N/A & 82.32 & 84.11 & 75.17 & 96.62 & 60.14 & 99.23 & 53.64 & 87.81 & 54.57 & 65.09 & 97.10 \\
        Hailuo & Diffusion & N/A & 83.41 & 84.85 & 77.65 & 97.05 & 64.91 & 99.22 & \underline{76.04} & 87.83 & 50.68 & \underline{75.50} & \underline{97.51} \\
        Sora & Diffusion & N/A & \underline{84.28} & \underline{85.51} & \underline{79.35} & 96.35 & \underline{79.91} & 98.74 & 70.85 & \underline{93.93} & \underline{56.95} & 74.29 & 96.23 \\
        ARLON $\ddagger$ & LLM+Diffusion & 1.5B & N/A & N/A & N/A & 97.10 & 52.77 & 98.92 & N/A & 89.80 & 54.43 & N/A & 93.41 \\
        ARLON $\dagger$ & LLM+Diffusion & 5B & 82.31 & 83.58 & 77.27 & \underline{98.65} & 72.22 & 97.56 & 74.49 & 90.60 & 52.07 & 62.53 & 95.59 \\
        \midrule
        \textbf{LanDiff} & LLM+Diffusion & 5B & \textbf{85.43} & \textbf{86.13} & \textbf{82.61} & \textbf{98.73} & \textbf{92.71} & 97.08 & \textbf{86.69} & \textbf{94.94} & 53.79 & 73.74 & 96.11 \\
        \bottomrule
    \end{tabular}
    }
\end{table*}

\begin{figure}[htbp]
    \centering
    \includegraphics[width=0.98\linewidth]{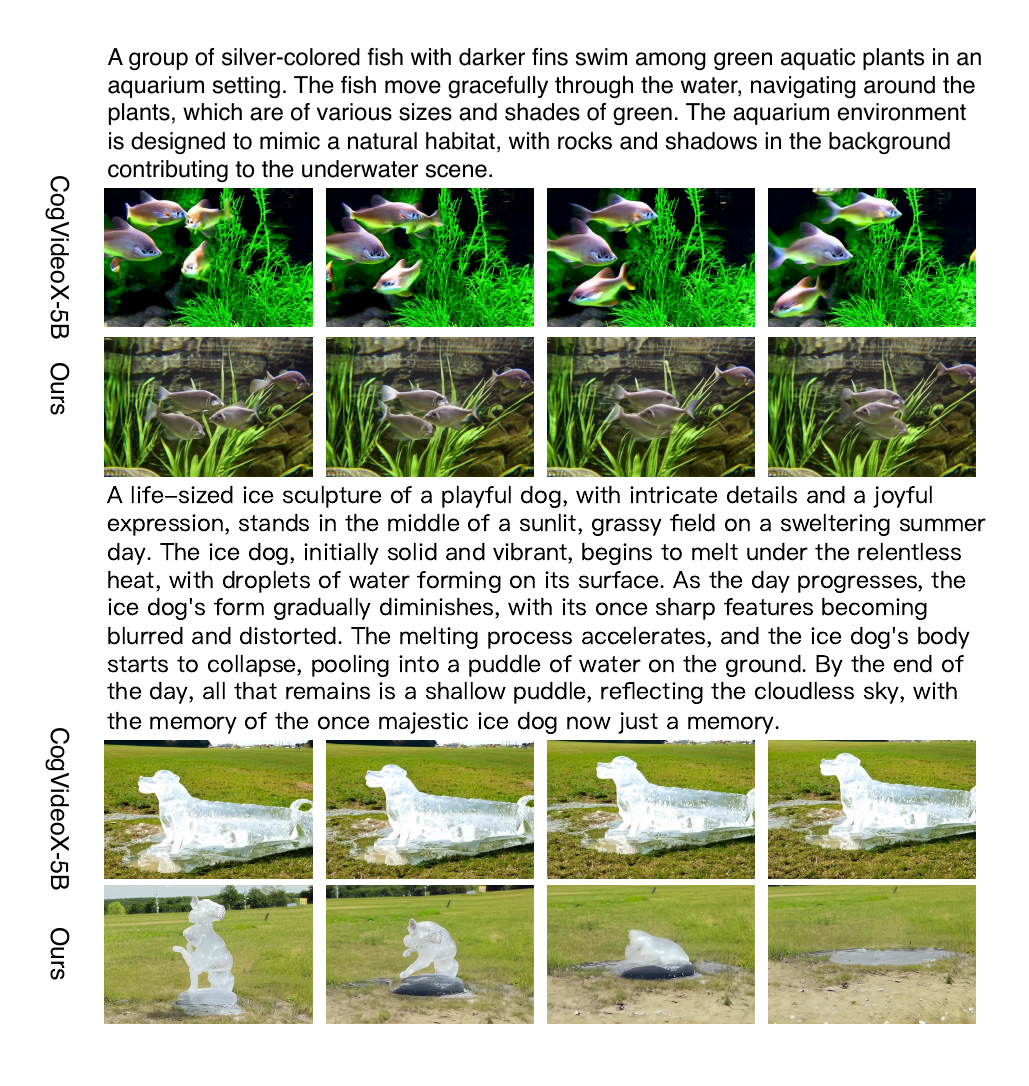}
    \caption{Comparison of qualitative results for text-to-video generation.}
    \label{fig:vis1}
\end{figure}

\textbf{Text to Video Generation.}
As shown in \autoref{tab:t2v}, we conducted a quantitative comparison between LanDiff and other state-of-the-art open-source models on the VBench benchmark.
Our model achieves the highest semantic score and quality score, indicating that our model can follow the text well and generate high-quality videos.
Compared with the same-sized CogVideoX-5B, our model achieves better results in almost all dimensions.
To eliminate the interference of the training data volume, we train a 7B DiT-based diffusion model with the same training data recipe as LanDiff. We mark it as DiT in the table.
This indicates that with the same data volume, by reasonably dividing the task load, we can combine the advantages of autoregressive models and diffusion models to achieve better performance than diffusion models alone.
\autoref{fig:vis1} shows the qualitative comparison between the videos generated by LanDiff and CogVideoX-5B.
In the first example, it can be seen that in the video generated by CogVideoX-5B, one fish disappears after the fish meet, while in the video generated by our model, the fish can still remain intact after meeting.
This demonstrates that our model can understand the concept of fish as an entity and maintain consistency over time.
Additionally, our model more faithfully adheres to the background elements described in the prompt ``with rocks and shadows in the background'', rendering these environmental details with greater accuracy and consistency.
In the second example, our model accurately captures the temporal dynamics described in the prompt, successfully rendering a melting ice sculpture of a dog. LanDiff properly depicts the progressive melting process, with the ice dog gradually losing its form and eventually transforming into a puddle of water. In contrast, CogVideoX-5B generates a static representation of the ice dog sculpture that remains largely unchanged throughout the sequence, failing to capture the crucial temporal narrative of melting described in the prompt. This highlights our model's superior capability in understanding and visualizing complex temporal transformations and physical processes.
For more video examples, please refer to our demo website\footnote{\label{fn:demo}\url{https://landiff.github.io/}}.

\begin{table*}[tbp]
    \centering
    \small
    \caption{Long video generation results of LanDiff  and other models on VBench. The best and second-best scores are highlighted in \textbf{bold} and \underline{underline}, respectively.}
        \resizebox{0.8\linewidth}{!}{

    \begin{tabular}{ccccccccc} %
        \toprule
    Models    & \thead{Total\\Score}      & \thead{Subject\\Consist}        & \thead{Background\\Consist}      & \thead{Motion\\Smooth}     & \thead{Dynamic\\Degree}    & \thead{Aesthetic \\Quality}  & \thead{Imaging \\Quality}   & \thead{Overall \\Consist}   \\  %
    \midrule
    FreeNoise & 62.60   & \underline{96.59}       & 97.48         & 98.36     & 17.44       & 47.39       &\textbf{63.88}    & 25.78      \\ %
    StreamingT2V & 62.92  & 87.31        & 94.64         & 93.83       &  \textbf{85.64}      & 44.57        & 53.64       & 23.65         \\  %
    OpenSora-V1.2 & 64.24   & 96.30         & 97.39          &\textbf{98.94}      & 44.79    & 56.68       & 51.64                &26.36    \\ %
    ARLON & \underline{65.09}   & \textbf{97.11}       & \underline{97.56}  & \underline{98.50}      & 50.42    & \underline{56.85}     & 53.85     &\underline{26.55} \\
    \midrule
    \textbf{LanDiff} & \textbf{68.34}   & 95.41       & \textbf{97.88}  & 97.38      & \underline{72.86}    & \textbf{60.96}     & \underline{63.00}     &\textbf{27.29} \\
    \bottomrule
    \end{tabular}
    }
    \label{tab:long}
    \end{table*}

    \begin{figure*}[htb!]
        \centering
        \includegraphics[width=0.7\linewidth]{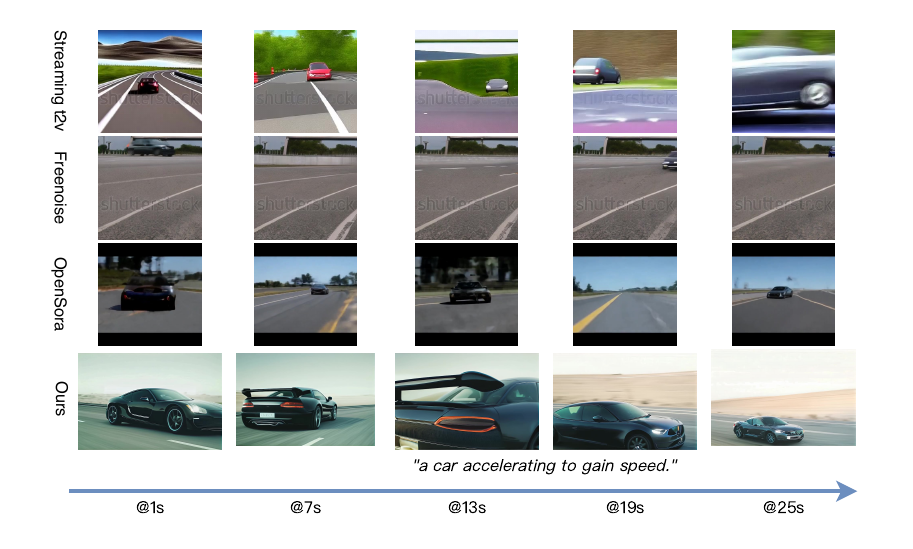}
        \caption{Qualitative comparison of long video generation results between LanDiff and other state-of-the-art models (FreeNoise, StreamingT2V and OpenSora-V1.2).}
        \label{fig:long_video_vis}
    \end{figure*}

\textbf{Long Video Generation.}
We compare our model with other open-source text-to-long video generation models: 1) FreeNoise \cite{qiu_FreeNoiseTuningfreeLonger_2024}. 2) StreamingT2V \cite{henschel_StreamingT2VConsistentDynamic_2024}. 3) OpenSora-V1.2 \cite{opensora}. 4) ARLON \cite{li_ARLONBoostingDiffusion_2024}.
As shown in \autoref{tab:long}, our LanDiff achieves state-of-the-art performance in long video generation tasks with the highest Total Score of 68.34, outperforming other open-source models across almost all dimensions. While StreamingT2V exhibits the highest Dynamic Degree (85.64), this comes at a clear cost to consistency metrics, as evidenced by its significantly lower Subject Consistency (87.31) and Background Consistency (94.64). An optimal video generation model should maximize dynamic content while maintaining temporal coherence. Compared with models that excel in consistency metrics but show limited dynamism (such as ARLON with 50.42 Dynamic Degree), our approach demonstrates superior dynamism (72.86) while preserving strong consistency scores and achieving the best results in Aesthetic Quality (60.96) and Overall Consistency (27.29). As illustrated in \autoref{fig:long_video_vis}, LanDiff effectively generates complex motion dynamics when prompted with ``a car accelerating to gain speed,'' successfully depicting the progressive increase in vehicle velocity while preserving visual consistency across frames. In contrast, FreeNoise, StreamingT2V, and OpenSora-V1.2 all struggle to properly render the acceleration motion, exhibiting either static vehicles, inconsistent car appearances, or unrealistic motion patterns that fail to convey the sense of increasing speed.
Additional video samples are available on our demo site\footref{fn:demo}.

\begin{figure}[htbp]
    \centering
    \includegraphics[width=0.98\linewidth]{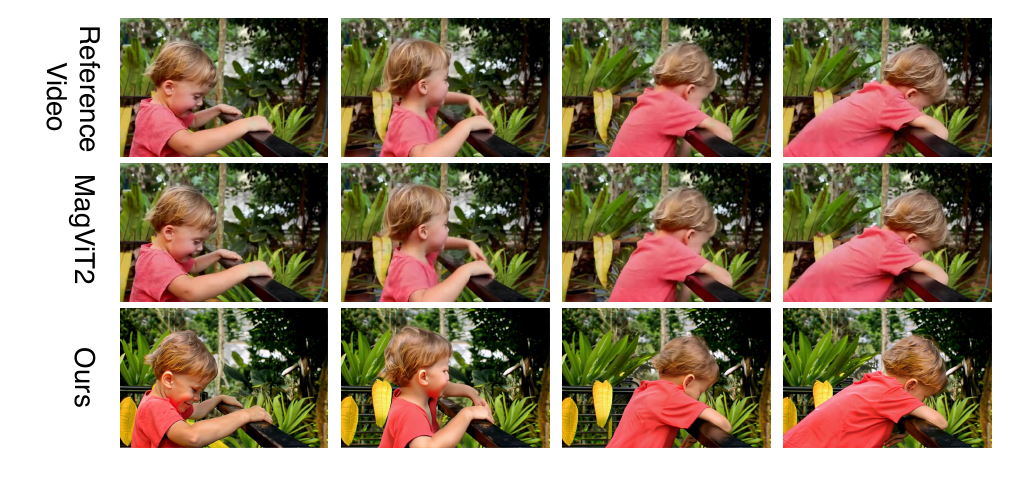}
    \caption{Visualization results of video reconstruction using video tokenizer.}
    \label{fig:vis_tokenizer}
\end{figure}

\begin{figure}[tbp!]
    \centering
    \includegraphics[width=0.98\linewidth]{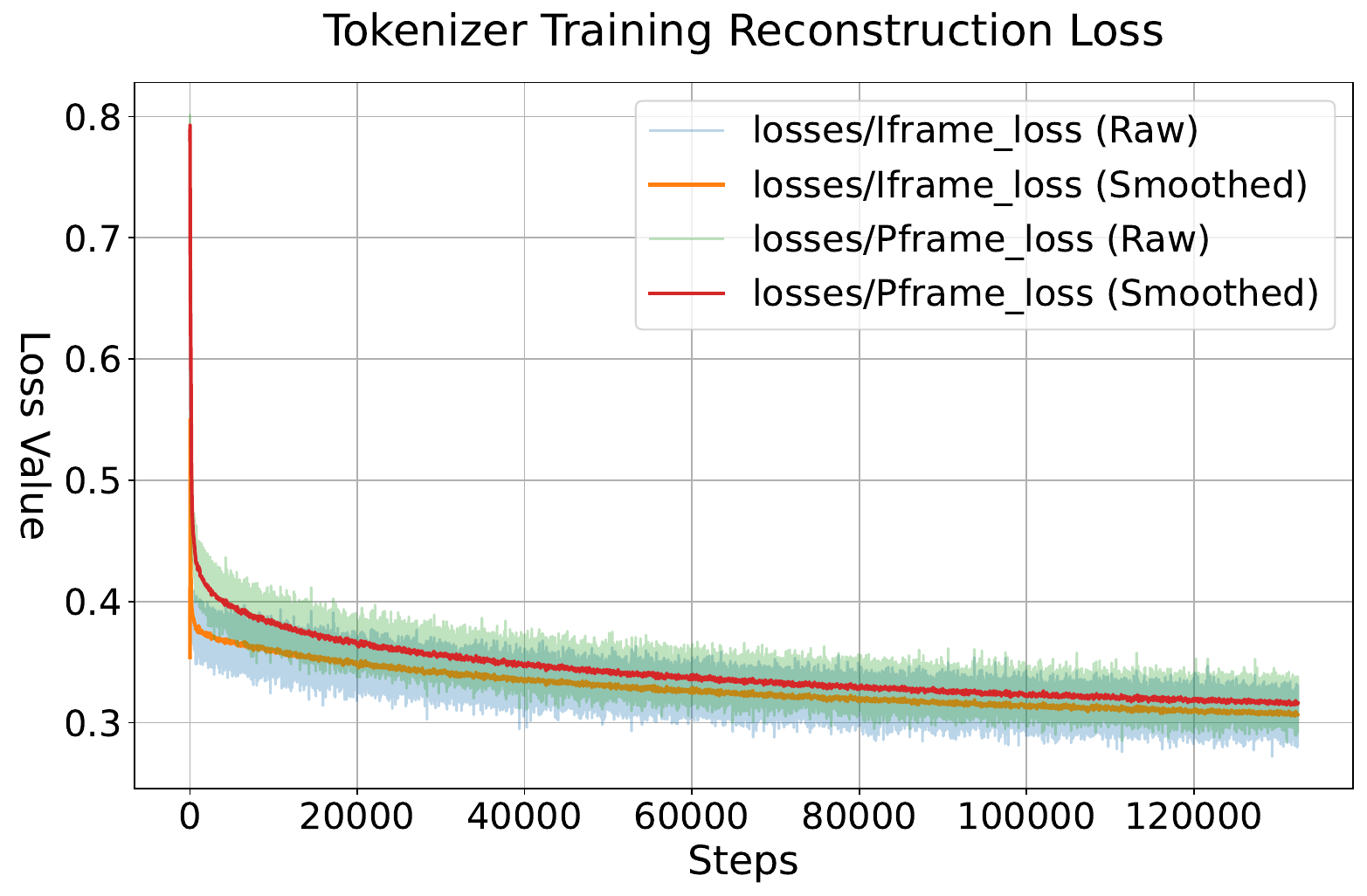}
    \vspace{-3mm}
    \caption{Training loss comparison of the video tokenizer. The plot illustrates the reconstruction loss trajectories for IFrame and PFrame components over training iterations. Despite the different token allocation strategies (330 tokens for IFrame vs. 74 tokens for PFrame), both frame types achieve comparable reconstruction quality.}
    \label{fig:tokenizer_loss}
    \vspace{-4mm}
\end{figure}

\textbf{Video Tokenization.}
We present visualization results of our model's video reconstruction in \autoref{fig:vis_tokenizer}.
We extract semantic tokens from the reference video and subsequently convert these tokens back into video using the video detokenizer.
The results demonstrate that through our careful design, our video tokenizer can accurately reconstruct videos with accurate semantics and actions using only approximately \textbf{1/50th} of the sequence length required by MagViT2 \cite{yu_LanguageModelBeats_2024}.
While some minor discrepancies in clothing details exist between the reconstructed and reference videos, these differences remain within acceptable limits for practical video generation applications.
Additional video samples are available on our demo site\footref{fn:demo}.
\autoref{fig:tokenizer_loss} shows the reconstruction loss trajectories for IFrame and PFrame components during training. Despite allocating significantly different token quantities (330 vs. 74), both frame types converge to comparable reconstruction quality. This validates our video frame grouping strategy that prioritizes key frames while minimizing tokens for intermediate frames. The results confirm our approach successfully balances reconstruction fidelity with computational efficiency.

\begin{table}[htbp]
    \centering
    \small
    \caption{The ablation study of video tokenizer and classifier free guidance (cfg) on Vbench benchmark.}
    \label{tab:ablation}
    \begin{tabular}{cccc}
        \toprule
    Models          & \thead{Total\\Score}        & \thead{Quality\\Score}      & \thead{Semantic\\Score}   \\  %
    \midrule
    LanDiff   & \textbf{85.43}       & \textbf{86.13}  & \textbf{82.61}   \\
    \midrule
    -video tokenizer   & 82.31       & 83.58  & 77.27   \\
    -cfg   & 81.06       & 83.04  & 73.13   \\
    \bottomrule
    \end{tabular}
    \vspace{-7mm}
    \end{table}

\subsection{Ablation Study}
In this section, we conducted ablation experiments on the video tokenizer and classifier-free guidance.
To fairly evaluate the effectiveness of our video tokenizer, we implemented an ARLON-like method as a comparison baseline. Specifically, for the control group without our proposed video tokenizer, we followed ARLON's architectural approach by training a large language model to predict quantized VAE features, then using CogVideoX-5B as the base diffusion model conditioned on these LLM-predicted VAE features to generate videos. This baseline model was trained with exactly the same dataset and training recipe as LanDiff, ensuring strict experimental control variables and effectively eliminating interference factors such as model parameter count and training data scale. As shown in \autoref{tab:ablation}, the experimental results clearly demonstrate that our proposed video tokenizer significantly improves both the quality score and semantic score of generated videos, validating the importance of this component in our framework.
In the case of not using classifier-free guidance for text, our model has decreased in quality and semantic scores.

%% file: sec/related_work.tex
\section{Related Work}

\textbf{Video Tokenization.}
Video tokenization plays a crucial role in video understanding and generation tasks.
Since a video can be represented as a sequence of continuous frames, some works directly use image tokenizers to process videos frame by frame.
For example, \citet{wang_Emu3NextTokenPrediction_2024} directly use SBER-MoVQGAN as the video tokenizer.
However, this method ignores the temporal redundancy in videos, resulting in a low compression rate.
To reduce temporal redundancy, some works \cite{yan_VideoGPTVideoGeneration_2021, yu_LanguageModelBeats_2024} try to extend image tokenizers based on 2D convolution to 3D convolution, which can process both temporal and spatial information simultaneously.
These methods encode videos in the original RGB space and perform video reconstruction tasks, which are more about perceptual compression.
In addition, some works \cite{ge_PlantingSEEDVision_2023,jin_UnifiedLanguageVisionPretraining_2023} try to train video tokenizers on features extracted from pre-trained visual encoders.
These tokenizers can achieve good performance in understanding and generation tasks while maintaining a high compression rate.
The video tokenizer we propose belongs to this category. Unlike previous feature-based tokenizers that achieved limited compression rates and primarily focused on image processing, our video tokenizer delivers significantly higher compression while handling both images and videos in a unified framework.

\textbf{LLM based Video Generation.}
LLM-based video generation methods are usually based on the Transformer \cite{vaswaniAttentionAllYou2017} structure, learning the mapping from text to video through next token prediction.
TATS \cite{ge_LongVideoGeneration_2022} uses VQ-GAN as the tokenizer and predicts video tokens using a GPT-like model structure.
VideoGPT \cite{yan_VideoGPTVideoGeneration_2021} uses 3D convolution to extract features and quantize them, and predicts the quantized video discrete tokens using a GPT-like model.
Recently, VideoPoet \cite{yu_LanguageModelBeats_2024} uses MagViT2 \cite{yu_LanguageModelBeats_2024} as the tokenizer and unifies multiple modalities as input to a large language model (LLM) to conditionally generate video tokens.
In addition, Emu3 \cite{wang_Emu3NextTokenPrediction_2024} uses SBER-MoVQGAN as the tokenizer to perform video understanding and generation by predicting the next token.
These works all use LLMs to directly generate ``perceptual features'' that contain rich visual details with high bit rates.
Recently, ARLON \cite{li_ARLONBoostingDiffusion_2024} attempts to discretize VAE features into a small number of tokens to reduce the bit rate required for LLMs' prediction.
In this way, the tokens retain low-frequency visual information such as blurry contours rather than high-level semantic information.
In contrast, our method employs tokens containing high-level semantic information as prediction targets for LLMs, which enables us to fully leverage LLMs' advantages in causal modeling to precisely generate high-quality videos.

\textbf{Diffusion based Video Generation.}
Diffusion-based methods have achieved great success in image generation, and recently many people \cite{chen_VideoCrafter2OvercomingData_2024, ho_ImagenVideoHigh_2022, ho_VideoDiffusionModels_2022, singer_MakeavideoTexttovideoGeneration_2023, zhou_MagicVideoEfficientVideo_2023} have tried to apply them to video generation tasks.
VDM \cite{ho_VideoDiffusionModels_2022} extends the 3D U-Net structure for video generation.
\citet{wang2024lavie} propose to generate high-quality and aesthetically pleasing videos in a cascaded manner.
Benefiting from the success of the text-to-image (T2I) field, some works such as Animatediff~\cite{guo_AnimateDiffAnimateYour_2024}, SVD~\cite{blattmann_StableVideoDiffusion_2023}, and PixelDance \cite{zeng_MakePixelsDance_2023} try to use pre-trained T2I models as initialization, and then add modules for temporal modeling to capture motion information for video generation.
\citet{ma_LatteLatentDiffusion_2024} explore the generation capabilities of multiple different structures of latent diffusion transformer.
After the release of SORA, a series of video generation methods based on the DiT \cite{peebles_ScalableDiffusionModels_2023} model have been proposed, including OpenSora~\cite{opensora}, OpenSoraPlan~\cite{lin_OpenSoraPlanOpenSource_2024}, Cogvideox~\cite{yang_CogVideoXTexttoVideoDiffusion_2024}, Hunyuan Video~\cite{kong_HunyuanVideoSystematicFramework_2024}, Mira~\cite{ju_MiraDataLargeScaleVideo_2024} and STIV~\cite{lin_STIVScalableText_2024} etc.
These methods can only generate short videos of a few seconds.
Recently, StreamingT2V \cite{henschel_StreamingT2VConsistentDynamic_2024} generates long videos by block-wise generation on a pre-trained short video generation model, and then uniformly performs mixed augmentation.
In addition, some works improve the consistency of long video generation by leveraging noise rescheduling techniques \cite{qiu_FreeNoiseTuningfreeLonger_2024,lu_FreeLongTrainingFreeLong_2024}.
Our method employs diffusion models as renderers for semantic features, enabling us to leverage their superior visual generation quality while circumventing their limitations in causal modeling.

%% file: sec/conclusion.tex
\section{Conclusion}

In this paper, we propose a new text-to-video generation model, LanDiff.
It combines the advantages of autoregressive models and diffusion models, including: 1) an efficient 1D video tokenizer to extract videos into semantic tokens; 2) a language model to generate semantic tokens based on text; 3) a video detokenizer to convert semantic tokens into videos.
LanDiff outperforms the state-of-the-art open-source models on the VBench benchmark, surpassing other open-source models in quality and semantic scores.
At the same time, LanDiff also achieves state-of-the-art performance in long video generation tasks.

\section{Broader Impact}

Text-to-video generation models such as LanDiff offer substantial potential for creative applications across entertainment, education, and content creation domains. Nevertheless, these technologies introduce ethical considerations and potential risks that warrant attention. The capability to generate photorealistic videos from textual descriptions could potentially be exploited to create misleading or deceptive content, including sophisticated deepfakes or videos that misrepresent individuals or events. Such misuse raises significant concerns regarding misinformation propagation, privacy violations, and potential harm to individuals or communities.
To mitigate these risks, we recommend several safeguards: (1) incorporating visible watermarks or robust digital signatures in generated content to ensure transparency regarding its synthetic nature; (2) advancing and deploying sophisticated detection systems capable of identifying AI-generated content with high accuracy; (3) establishing comprehensive usage policies that explicitly prohibit harmful applications; and (4) implementing accessible reporting mechanisms for suspected misuse cases. Furthermore, we advocate for continued research into technical safeguards and responsible deployment frameworks specifically designed for generative video models.
We emphasize that our research contribution aims to advance the field of multimodal generation for socially beneficial applications while acknowledging the necessity of addressing potential negative impacts through complementary technical innovations and policy measures.

\section{Limitation and Future Works}

While LanDiff demonstrates significant advancements in text-to-video generation, several limitations remain to be addressed in future work. First, the scale of our language model (2B parameters) is substantially smaller than state-of-the-art text-only LLMs, potentially limiting the semantic understanding and generation capabilities. Future work will explore scaling our language models to larger parameter counts to enhance performance.
Second, our choice of CogVideoX-2B as the underlying diffusion model establishes an upper bound on the quality of generated videos. We plan to investigate the development of more sophisticated diffusion backbones specifically optimized for video generation tasks to overcome this constraint.
Third, we observe that LanDiff struggles with accurate text rendering within generated videos. This limitation likely stems from insufficient supervision of text features in the current semantic token representation. Future research will focus on developing more comprehensive video semantic tokens with enhanced text-specific supervision.
Our current work primarily addresses text-to-video generation, but we envision extending LanDiff's capabilities to broader application scenarios. These include image-to-video generation, unified models for video understanding and generation, and interactive controllable video synthesis. These extensions would significantly expand the utility of semantic token-based language models in multimodal generation tasks and provide more flexible creative tools for users across various domains.

%% file: sec/appendix.tex
\section{Implementation Details}

\begin{table}[thbp!]
    \centering
    \small
    \caption{The detailed model configurations of Video Tokenizer.}
    \label{tb:tokenizer_config_details}
    \begin{tabular}{cccc }
      \toprule
      Module     & Configuration    & Value & \#Parameters \\
      \midrule
      \multirow{5}{*}{Encoder} &  Transformer Layer & 12  & \multirow{5}{*}{85M}\\
      & Hidden Size & 768 &\\
      & Attention Heads & 12& \\
      &  IFrame Query Tokens & 330& \\
      &  PFrame Query Tokens & 74& \\
      \midrule
      \multirow{4}{*}{Decoder} &  Transformer Layer & 12  & \multirow{4}{*}{85M}\\
      & Hidden Size & 768 &\\
      & Attention Heads & 12& \\
      & Mask Tokens & 1& \\
      \midrule
      \multirow{3}{*}{Quantizer} &  Codebook size & 2048 & \multirow{3}{*}{25.4K}\\
      & Codebook Dimension & 16 & \\
      &  Similarity Metric & Cosine& \\
      \midrule
      \multicolumn{3}{c}{Total Parameters} & 170M \\    
      \bottomrule
    \end{tabular}
  \end{table}
  \begin{table}[htb!]
    \centering
    \caption{The detailed model configurations of LanDiff.}
    \label{fig:model_config_details}
    \begin{tabular}{cccc }
      \toprule
      Module     & Configuration    & Value & \#Parameters \\
      \midrule
      \multirow{7}{*}{LLM} &  Transformer Layer & 24  & \multirow{7}{*}{2B}\\
      & Hidden Size & 2048 &\\
      & Attention Heads & 16& \\
      & MLP Dimension & 11008 &\\
      & Activation & GELU &\\
      & RoPE $\theta$ & 10000 &\\
      & Text Drop Rate & 0.1 &\\
      & Micro Conditioner Hidden Size & 512 &\\
      \midrule
      \multirow{4}{*}{Diffusion Backbone Module (Frozen)} &  Transformer Layer & 30 & \multirow{4}{*}{2B} \\
      & Attention Heads & 8 & \\
      & Hidden Size & 1920 &\\
      & Time Embedding Size & 256& \\ 
      \midrule
      \multirow{4}{*}{Diffusion Control Module} &  Transformer Layer & 15 & \multirow{4}{*}{1B} \\
      & Attention Heads & 8 & \\
      & Hidden Size & 1920 &\\
      & Time Embedding Size & 256& \\ 
      \midrule
      \multicolumn{3}{c}{Trainable Parameters} & 3B \\  
      \midrule  
      \multicolumn{3}{c}{Total Parameters} & 5B \\    
      \bottomrule
    \end{tabular}
  \end{table}

\subsection{Details of Video Tokenizer}
The video tokenizer model follows a similar structure to TiTok \cite{yu_ImageWorth32_2024}.
However, to flexibly encode videos with different numbers of frames, we replace absolute position encoding with 3D RoPE position encoding \cite{su_RoFormerEnhancedTransformer_2024}.
The model configuration of the tokenizer and the parameters of each part are shown in \autoref{tb:tokenizer_config_details}.
The model is a Transformer structure, with 12 layers in both the encoder and decoder, a hidden layer size of 768, and 12 heads.
To improve the computational efficiency of the attention mechanism in the tokenizer, we employ flex attention\cite{dong_FlexAttentionProgramming_2024}.
In addition, inspired by EnCodec \cite{defossez_HighFidelityNeural_2022}, to avoid discontinuities when encoding videos, we set a 20\% overlap between groups.
During training, the batch size is 96, we use the AdamW \cite{loshchilov2017decoupled} optimizer, the learning rate is constant at $1e-4$, and the learning rate decay factor is 0.
During training, we use Model Exponential Moving Average (EMA) to smooth the model parameters, and the decay rate of EMA is 0.8.
The weights of the reconstruction loss and the commitment loss are both 1.

\subsection{Details of LLM}
We use a model structure similar to LLaMA \cite{touvron_LLaMAOpenEfficient_2023} as the LLM.
The model has 24 layers, a hidden layer size of 2048, 16 heads, and an MLP hidden layer size of 11008.
The batch size for model training is 4096, we use the AdamW optimizer, the learning rate is $1e-3$, the learning rate decay factor is 0.1, and we use a warm-up strategy for the first 1000 steps of training.
We use a cosine learning rate decay strategy.

\subsection{Details of Diffusion Model}
The batch size for training is 128, we use the AdamW optimizer, the learning rate is $1e-4$, and the learning rate decay factor is $1e-4$.
To speed up training, we first train directly on the original features extracted from Theia.
Then we use the quantized reconstructed features for training.

\begin{table*}[htbp]
  \caption{\textbf{Performance comparison of Text-to-video (T2V) generation between our LanDiff and other state-of-the-art models on VBench benchmark.} The remaining 8 evaluation dimensions of VBench that are not shown in the main text. The best and second-best scores are highlighted in \textbf{bold} and \underline{underline}, respectively. $\dagger$ indicates the scores we reproduced, while $\ddagger$ indicates the scores from the original papers, and other scores are from the VBench benchmark.}
  \label{tab:t2v_left}
  \small
  \setlength\tabcolsep{2pt}
  \centering
  \resizebox{0.9\linewidth}{!}{
  \begin{tabular}{lccccccccccccc}
      \toprule
      \thead{\textbf{Model} \\ \textbf{Name}} & \thead{\textbf{Type}} & \thead{\textbf{Model} \\ \textbf{Size}} & \thead{\textbf{Total} \\ \textbf{Score}} & \thead{\textbf{Quality} \\ \textbf{Score}} & \thead{\textbf{Semantic} \\ \textbf{Score}} & \thead{\textbf{aesthetic} \\ \textbf{quality}} & \thead{\textbf{appearance} \\ \textbf{style}} & \thead{\textbf{color}} & \thead{\textbf{human} \\ \textbf{action}} & \thead{\textbf{imaging} \\ \textbf{quality}} & \thead{\textbf{overall} \\ \textbf{consistency}} & \thead{\textbf{temporal} \\ \textbf{flickering}} & \thead{\textbf{temporal} \\ \textbf{style}} \\
      \midrule
      \multicolumn{12}{c}{\textit{Open Sourced Models}} \\
      \midrule
      InstructVideo & Diffusion & 1.3B & 76.61 & 81.56 & 56.81 & 52.55 & 20.16 & 77.14 & 85.20 & \underline{68.01} & 19.91 & 98.19 & 21.26 \\
      Latte-1 & Diffusion & 0.7B & 77.29 & 79.72 & 67.58 & 61.59 & 23.74 & 85.31 & 90.00 & 61.92 & 27.33 & 98.89 & 24.76 \\
      OpenSoraPlan V1.1 & Diffusion & 2.7B & 78.00 & 80.91 & 66.38 & 56.85 & 22.90 & 89.19 & 86.80 & 62.28 & 26.52 & 99.03 & 23.87 \\
      Show-1 & Diffusion & 6.3B & 78.93 & 80.42 & 72.98 & 57.35 & 23.06 & 86.35 & 95.60 & 58.66 & 27.46 & 99.12 & 25.28 \\
      OpenSora V1.2 & Diffusion & 1.1B & 79.76 & 81.35 & 73.39 & 56.85 & 23.95 & 90.08 & 91.20 & 63.34 & 26.85 & \underline{99.53} & 24.54 \\
      LTX-Video & Diffusion & 1.9B & 80.00 & 82.30 & 70.79 & 59.81 & 21.47 & 81.45 & 92.80 & 60.28 & 25.19 & 99.34 & 22.62 \\
      Mochi-1 & Diffusion & 10B & 80.13 & 82.64 & 70.08 & 56.94 & 20.33 & 79.73 & 94.60 & 60.64 & 25.15 & 99.40 & 23.65 \\
      AnimateDiff-V2 & Diffusion & 1.3B & 80.27 & 82.90 & 69.75 & \underline{67.16} & 22.42 & 87.47 & 92.60 & 70.10 & 27.04 & 98.75 & \textbf{26.03} \\
      VideoCrafter-2.0 & Diffusion & 1.7B & 80.44 & 82.20 & 73.42 & 63.13 & 25.13 & \textbf{92.92} & 95.00 & 67.22 & \textbf{28.23} & 98.41 & \underline{25.84} \\
      CogVideoX-2B & Diffusion & 2B & 80.91 & 82.18 & 75.83 & 60.82 & 24.80 & 79.41 & 98.00 & 61.68 & 26.66 & 98.89 & 24.36 \\
      Emu3 $\ddagger$ & LLM & 8B & 80.96 & N/A & N/A & 59.64 & 20.92 & N/A & 77.71 & N/A & N/A & N/A & N/A \\
      Vchitect-2.0-2B & Diffusion & 2B & 81.57 & 82.51 & 77.79 & 61.47 & 24.93 & 86.87 & 97.00 & 65.60 & \underline{28.01} & 98.45 & 25.56 \\
      CogVideoX-5B & Diffusion & 5B & 81.61 & 82.75 & 77.04 & 61.98 & 24.91 & 82.81 & \textbf{99.40} & 62.90 & 27.59 & 98.66 & 25.38 \\
      DiT $\dagger$ & Diffusion & 7B & 81.85 & 82.70 & 78.42 & 60.00 & 24.95 & 78.62 & 98.20 & 63.80 & 27.85 & 99.13 & 26.10 \\
      RepVideo & Diffusion & 2B & 81.94 & 82.70 & 78.91 & 62.40 & 25.12 & 82.51 & 98.00 & 63.16 & 26.96 & 99.16 & 25.31 \\
      Vchitect-2.0[E] & Diffusion & 2B & 82.24 & 83.54 & 77.06 & 60.41 & 23.73 & 87.04 & 97.20 & 65.35 & 27.57 & 98.57 & 25.01 \\
      HunyuanVideo & Diffusion & 13B & 83.24 & 85.09 & 75.82 & 60.36 & 19.80 & \underline{91.60} & 94.40 & 67.56 & 26.44 & 99.44 & 23.89 \\
      \midrule
      \multicolumn{12}{c}{\textit{Close Sourced Models}} \\
      \midrule
      Pika-1.0 & Diffusion & N/A & 80.69 & 82.92 & 71.77 & 62.04 & 22.26 & 90.57 & 86.20 & 61.87 & 25.94 & \textbf{99.74} & 24.22 \\
      Kling & Diffusion & N/A & 81.85 & 83.39 & 75.68 & 61.21 & 19.62 & 89.90 & 93.40 & 65.62 & 26.42 & 99.30 & 24.17 \\
      Jimeng & Diffusion & N/A & 81.97 & 83.29 & 76.69 & \textbf{68.80} & 22.27 & 89.05 & 90.10 & 67.09 & 27.10 & 99.03 & 24.70 \\
      Gen-3 & Diffusion & N/A & 82.32 & 84.11 & 75.17 & 63.34 & 24.31 & 80.90 & 96.40 & 66.82 & 26.69 & 98.61 & 24.71 \\
      Hailuo & Diffusion & N/A & 83.41 & 84.85 & 77.65 & 63.03 & 20.06 & 90.36 & 92.40 & 67.17 & 27.10 & 99.10 & 25.63 \\
      Sora & Diffusion & N/A & \underline{84.28} & \underline{85.51} & \underline{79.35} & 63.46 & 24.76 & 80.11 & \underline{98.20} & \textbf{68.28} & 26.26 & 98.87 & 25.01 \\
      ARLON $\ddagger$ & LLM+Diffusion & 1.5B & N/A & N/A & N/A & 61.01 & N/A & N/A & N/A & 60.98 & 27.27 & 99.37 & 25.33 \\
      ARLON $\dagger$ & LLM+Diffusion & 5B & 82.31 & 83.58 & 77.27 & 60.58 & \underline{25.26} & 85.86 & 92.20 & 62.72 & 26.14 & 99.39 & 24.07 \\
      \midrule
      \textbf{LanDiff} & LLM+Diffusion & 5B & \textbf{85.43} & \textbf{86.13} & \textbf{82.61} & 64.78 & \textbf{25.60} & 91.09 & 97.20 & 65.69 & 27.43 & 99.43 & 25.26 \\
      \bottomrule
  \end{tabular}
  }
\end{table*}
\begin{figure}[htbp]
  \centering
  \includegraphics[width=0.7\linewidth]{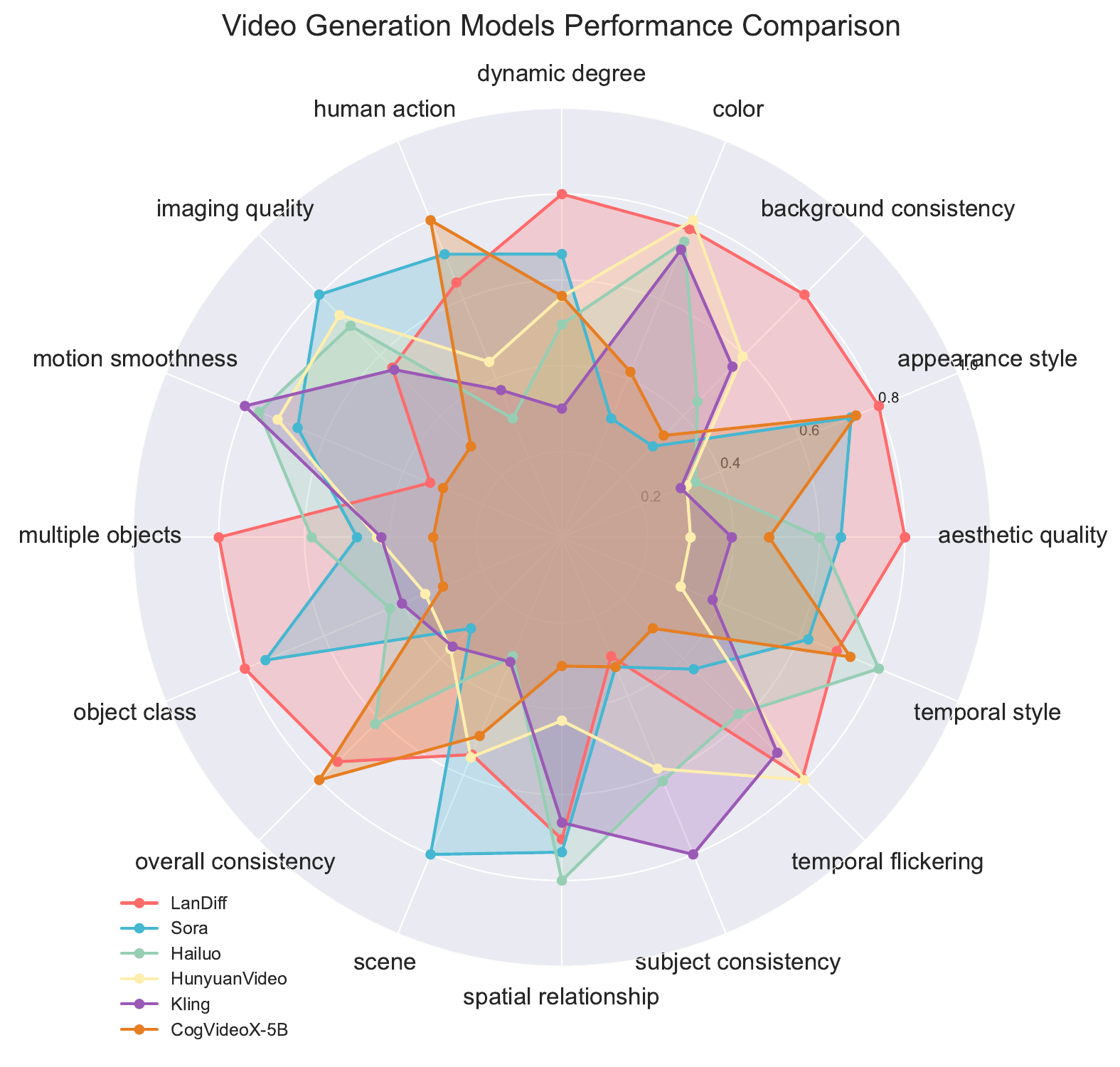}
  \caption{\textbf{Radar chart visualization of performance comparison across different dimensions on VBench.} The plot compares LanDiff against five competitive baselines: Sora, Hailuo, HunyuanVideo, Kling, and CogVideoX-5B. For better readability, the values in the radar chart have been normalized to a scale ranging from 0.3 to 0.8. The normalization was performed using the min-max scaling formula: $normalized = 0.3 + 0.5 \times \frac{value - min\_value}{max\_value - min\_value}$. The original raw performance data can be found in \autoref{tab:t2v} and \autoref{tab:t2v_left}.}
  \label{fig:radar_chart}
\end{figure}

\section{More Analysis on VBench Benchmark}

As shown in \autoref{tab:t2v_left}, we conduct a comprehensive comparison with state-of-the-art text-to-video generation models on VBench benchmark. The benchmark evaluates models across multiple dimensions including quality, semantics, aesthetics, and temporal consistency. Our LanDiff achieves superior performance in most metrics, particularly excelling in overall quality (86.13) and semantic accuracy (82.61).

Among open-sourced models, there is a clear trend of performance improvement with model size, from Latte-1 (0.7B) to HunyuanVideo (13B). However, our hybrid LLM+Diffusion approach (5B) demonstrates that architectural innovation can be more impactful than simply scaling up model parameters. Notably, LanDiff outperforms much larger models like HunyuanVideo (13B) and Mochi-1 (10B) across most metrics.

As visualized in \autoref{fig:radar_chart}, we compare LanDiff with five representative models across different evaluation dimensions. The radar chart reveals that LanDiff (shown in red) demonstrates well-balanced performance across all metrics, with notably strong results in quality score and semantic accuracy. While Sora (shown in blue) achieves competitive scores in imaging quality and scene, and HunyuanVideo excels in certain visual aspects, LanDiff maintains consistently superior performance across the entire spectrum of metrics. 
Notably, while there is typically a trade-off between dynamic degree and motion smoothness/subject consistency, LanDiff achieves a high level of dynamism while maintaining strong performance in stability metrics - with subject consistency and motion smoothness scores within 2.3\% and 2.5\% of the best-performing models respectively.
Notably, while text-to-video models typically exhibit a trade-off between dynamic expressiveness and temporal stability, LanDiff successfully balances these competing objectives—achieving high dynamism while maintaining robust stability metrics, with subject consistency and motion smoothness scores deviating by only 2.3\% and 2.5\% respectively from the state-of-the-art in each category.
The comprehensive comparison with these strong baselines, including both commercial (Sora, Hailuo, Kling) and open-source models (HunyuanVideo, CogVideoX-5B), further validates the effectiveness of our hybrid LLM+Diffusion approach.

\section{More Examples}
\begin{figure*}[htbp]
  \centering
  \includegraphics[width=0.9\linewidth]{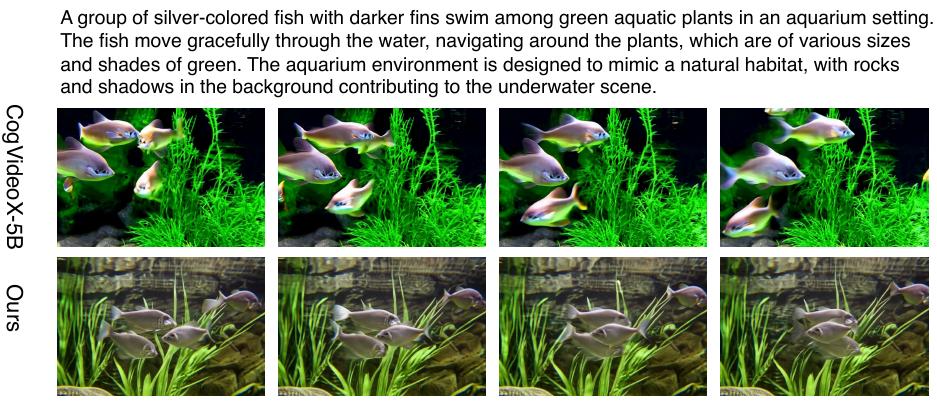}
  \caption{Examples of text to videos generation of LanDiff and CogVideoX-5B.}
\end{figure*}
\begin{figure*}[htbp]
  \centering
  \includegraphics[width=0.9\linewidth]{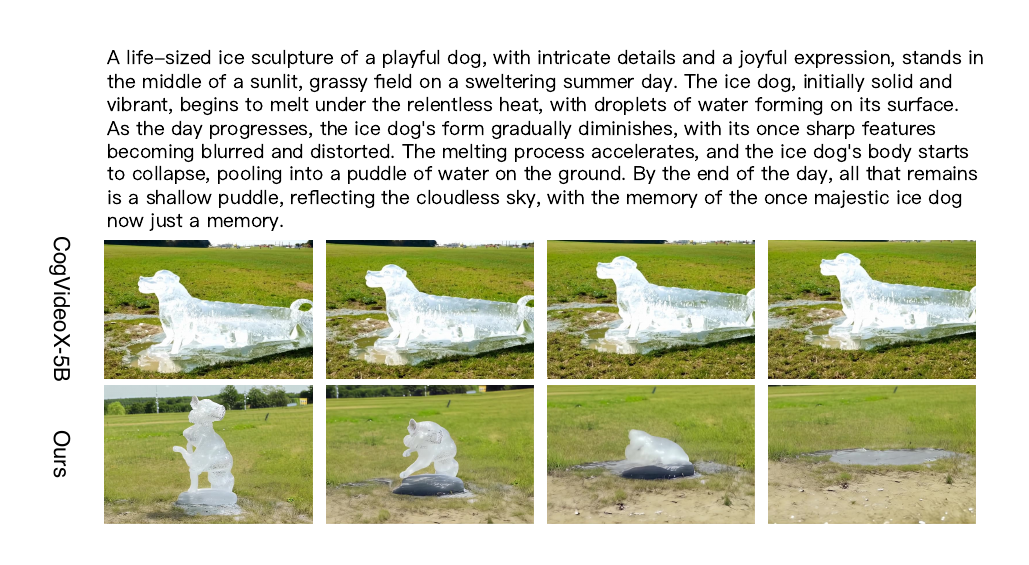}
  \caption{Examples of text to videos generation of LanDiff and CogVideoX-5B.}
\end{figure*}
\begin{figure*}[htbp]
  \centering
  \includegraphics[width=0.9\linewidth]{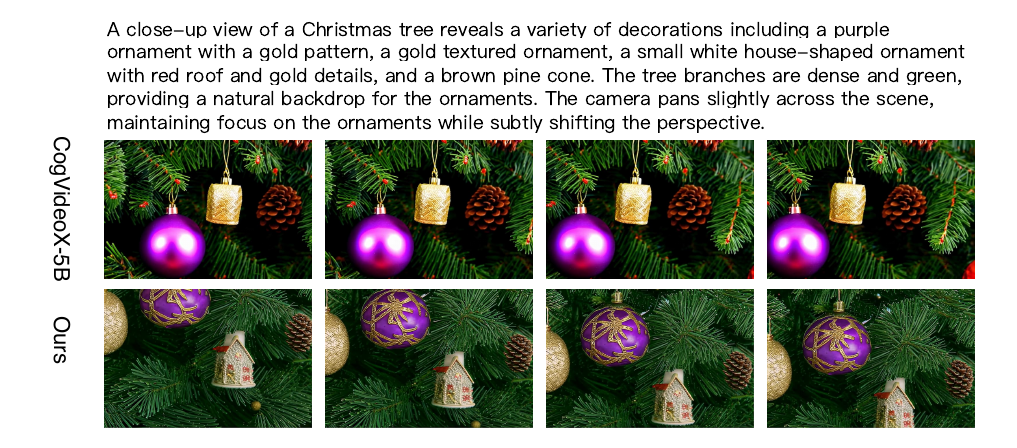}
  \caption{Examples of text to videos generation of LanDiff and CogVideoX-5B.}
\end{figure*}
\begin{figure*}[htbp]
  \centering
  \includegraphics[width=0.9\linewidth]{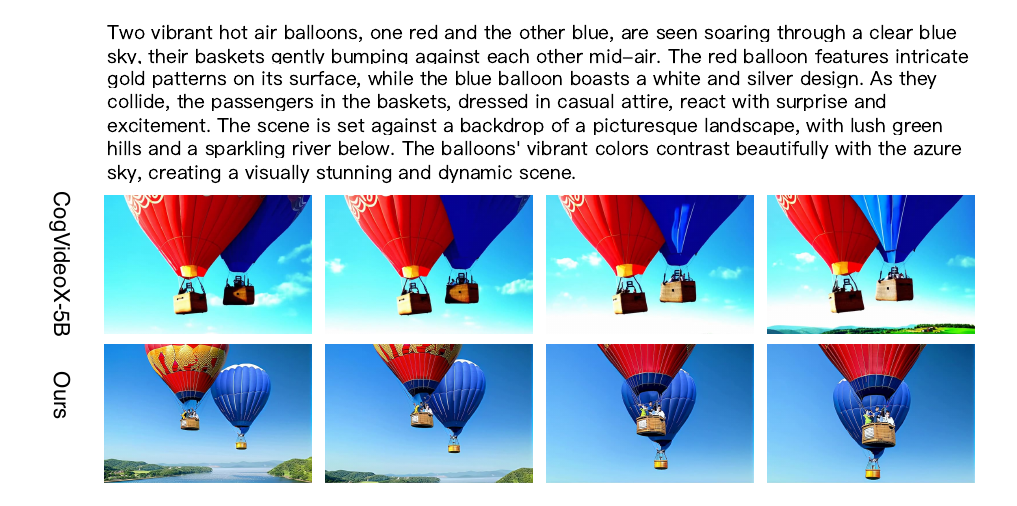}
  \caption{Examples of text to videos generation of LanDiff and CogVideoX-5B.}
\end{figure*}
\begin{figure*}[htbp]
  \centering
  \includegraphics[width=0.9\linewidth]{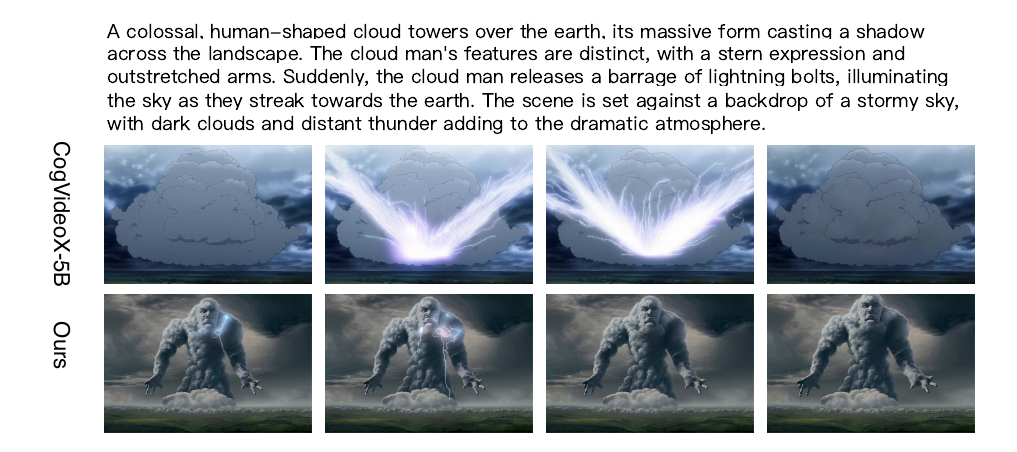}
  \caption{Examples of text to videos generation of LanDiff and CogVideoX-5B.}
\end{figure*}
\begin{figure*}[htbp]
  \centering
  \includegraphics[width=0.9\linewidth]{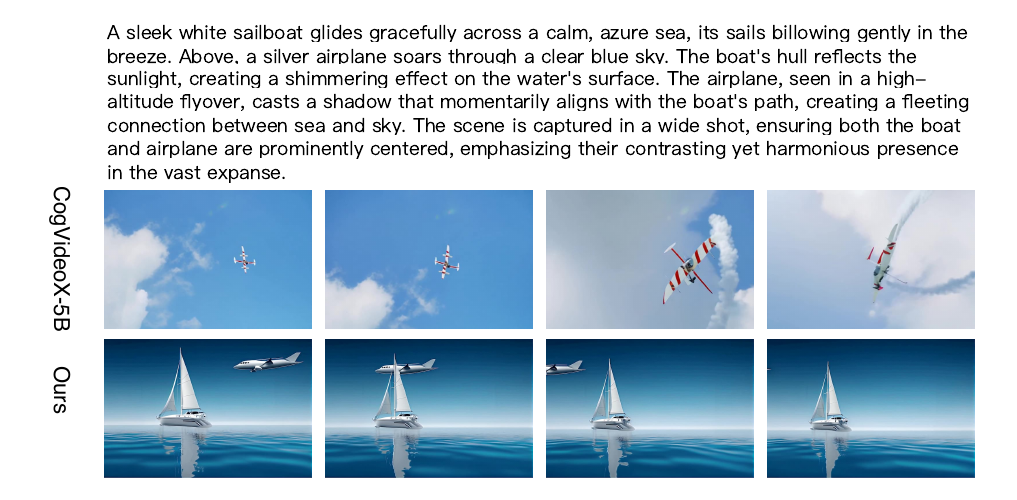}
  \caption{Examples of text to videos generation of LanDiff and CogVideoX-5B.}
\end{figure*}

\begin{figure*}[htbp]
  \centering
  \includegraphics[width=0.9\linewidth]{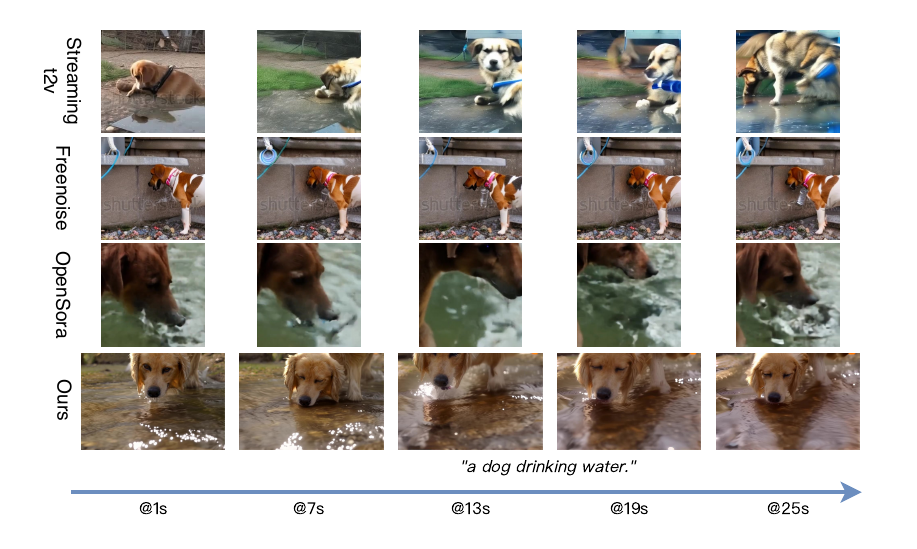}
  \caption{Examples of text to long video generation.}
\end{figure*}

\begin{figure*}[htbp]
  \centering
  \includegraphics[width=0.9\linewidth]{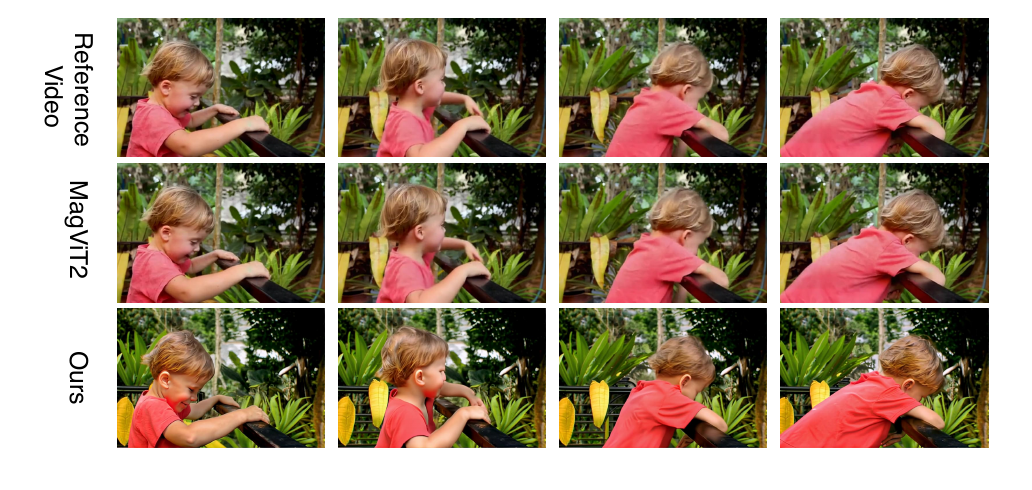}
  \caption{Visualization results of video reconstruction using video
  tokenizer.}
\end{figure*}

\begin{figure*}[htbp]
  \centering
  \includegraphics[width=0.9\linewidth]{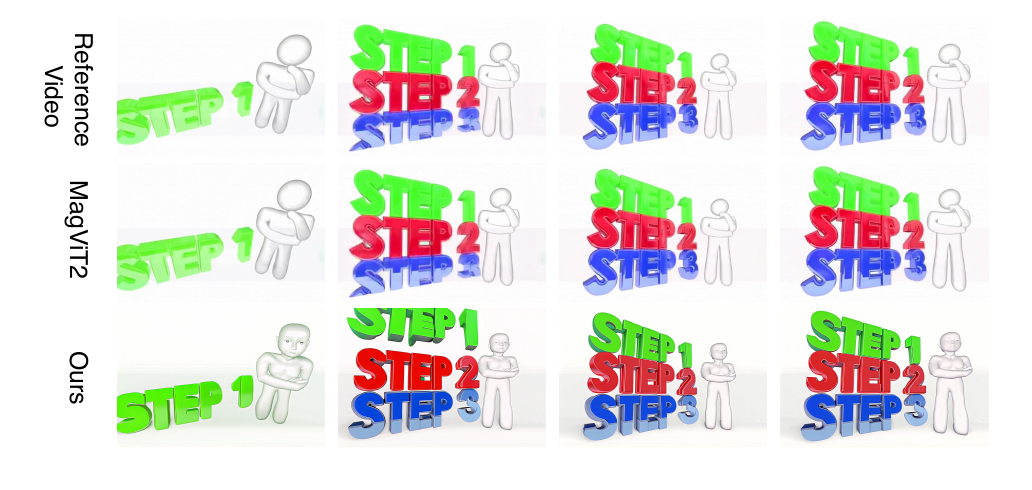}
  \caption{Visualization results of video reconstruction using video
  tokenizer.}
\end{figure*}